\documentclass{article}
\PassOptionsToPackage{numbers, compress}{natbib}

\usepackage[main,final]{neurips_2026}

\usepackage[utf8]{inputenc}
\usepackage[T1]{fontenc}
\usepackage{hyperref}
\usepackage{url}
\usepackage{booktabs}
\usepackage{amsfonts}
\usepackage{nicefrac}
\usepackage{microtype}
\usepackage[table]{xcolor}

\usepackage{graphicx}
\usepackage{subcaption}
\usepackage{xspace}
\usepackage{amsmath}
\usepackage{amssymb}
\usepackage{mathtools}
\usepackage{amsthm}
\usepackage{multirow}
\usepackage{float}
\usepackage{algorithm}
\usepackage{algpseudocode}
\usepackage{wrapfig}
\usepackage{placeins}

\usepackage[capitalize,noabbrev]{cleveref}

\theoremstyle{plain}

\theoremstyle{definition}

\theoremstyle{remark}

\usepackage[disable,textsize=tiny]{todonotes}

\newcommand{\name}{Step-dLLM\xspace}

\title{\name: Adaptive Step-aware Sparse Attention for Efficient Diffusion LLM Inference}

\author{%
  \textbf{Zhichen Zeng}\textsuperscript{1}\thanks{Equal Contribution.}\hspace{4pt}
  \quad \textbf{Xichong Zhang}\textsuperscript{2}\footnotemark[1]
  \quad \textbf{Junpan Wu}\textsuperscript{3}
  \quad \textbf{Yifei Zuo}\textsuperscript{4}
  \quad \textbf{Chi-Chih Chang}\textsuperscript{5}\\
  \textbf{Jiayi Wang}\textsuperscript{1}
  \quad \textbf{Maohua Nie}\textsuperscript{1}
  \quad \textbf{Ji Liu}\textsuperscript{6}
  \quad \textbf{Ang Li}\textsuperscript{1}
  \quad \textbf{Banghua Zhu}\textsuperscript{1}\\[3pt]
  \textsuperscript{1}University of Washington
  \quad \textsuperscript{2}University of Waterloo
  \quad \textsuperscript{3}University of Wisconsin, Madison\\
  \textsuperscript{4}Northwestern University
  \quad \textsuperscript{5}Cornell University
  \quad \textsuperscript{6}Meta AI
}

\begin{document}

\raggedbottom

\maketitle
\begin{abstract}
Diffusion large language models (dLLMs) generate tokens in parallel via iterative denoising, but their bidirectional attention prevents the KV caching used by autoregressive models and recomputes the full attention map at every step, incurring quadratic per-step cost. Existing sparse attention methods for dLLMs construct a sparse mask after an initial full-attention warm-up and reuse it for the remaining steps, while allocating the sparsity budget uniformly across heads. We argue both choices ignore key structural properties of attention in dLLMs: empirically, attention patterns are only locally consistent across steps and shift abruptly at certain transitions, and different heads concentrate their attention mass to very different extents. To this end, we design \textbf{Step-dLLM}, which periodically refreshes a block-sparse mask at anchor steps along the denoising trajectory and reuses it within each locally consistent interval, while employing a head-adaptive global budget that more aggressively compresses heads with concentrated attention. Backed by a fused IO-aware Triton kernel for block-sparse attention, Step-dLLM applies to pre-trained LLaDA-1.5 and Dream-7B-Instruct without retraining, achieving a $2.78\times$ end-to-end speedup over dense FlashAttention on LLaDA-1.5 at 8K context with 256 denoising steps and $90\%$ sparsity, while improving the accuracy--latency trade-off over prior sparse-attention methods.

\end{abstract}

\section{Introduction}
\label{sec:intro}

Recently, diffusion large language models (dLLMs) have achieved significant progress in the area of natural language processing~\cite{nie2025llada,llada-moe2025,bie2025llada2,arriola2025block,ye2025dream,song2025seeddiffusion}. Traditional autoregressive language models (ARs), such as DeepSeek~\cite{deepseek_v32} and Qwen~\cite{qwen_qwen36_max}, rely on causal attention mechanisms, which restrict each token to attending only to its preceding context. In contrast, dLLMs leverage bidirectional context understanding and generate the entire sequence in parallel. By exploiting this capability, dLLMs achieve strong performance in language generation and represent a promising alternative to ARs.

Despite the advantages of parallel decoding, dLLMs still suffer from high generation latency. In AR models, KV caching~\cite{zhang2023h2o,li2024snapkv} allows each decoding step to reuse previously computed key--value pairs, reducing the per-step attention cost to linear in the sequence length. In contrast, dLLMs~\cite{nie2025llada,bie2025llada2,ye2025dream} rely on bidirectional attention, where all tokens, including those under denoising, attend to one another. Since token representations are updated at every denoising step, exact reuse of the standard autoregressive KV cache is unavailable without additional assumptions, and vanilla dLLMs recompute full-sequence attention repeatedly. As a result, dLLMs incur quadratic per-step cost across multiple denoising iterations, which limits their practical efficiency on long contexts~\cite{wu2025fastdllm,d2f2025,dinfer2025}.

To improve dLLM efficiency, recent work has explored two complementary directions. The first is semi-autoregressive generation, exemplified by block diffusion~\cite{wu2025fastdllm}, which partitions the output into blocks, applies diffusion within each block, and decodes blocks autoregressively. This shortens each diffusion pass and enables KV caching for previous blocks. The second direction is sparse attention, which is orthogonal to block diffusion and reduces cost by attending only to salient token interactions within each step. This paper focuses on sparse attention. Although sparse attention has been widely studied in AR models~\cite{jiang2024minference, xiao2023efficient}, those methods rely on structured causal patterns and do not directly transfer to dLLMs, where bidirectional attention evolves as tokens are progressively denoised. SparseD~\cite{wang2025sparsed} takes a first step toward sparse attention for dLLMs that runs an initial fraction of denoising steps with full attention, constructs a sparse mask from these attention maps, and reuses it for the remaining steps. This exploits substantial cross-step similarity, but our measurements show that high similarity within a local interval does not guarantee that all salient connections persist across distant steps.

\begin{wrapfigure}[23]{r}{0.36\columnwidth}
    \centering
    \vspace{-1em}
    \begin{subfigure}[b]{\linewidth}
        \centering
        \includegraphics[width=\linewidth]{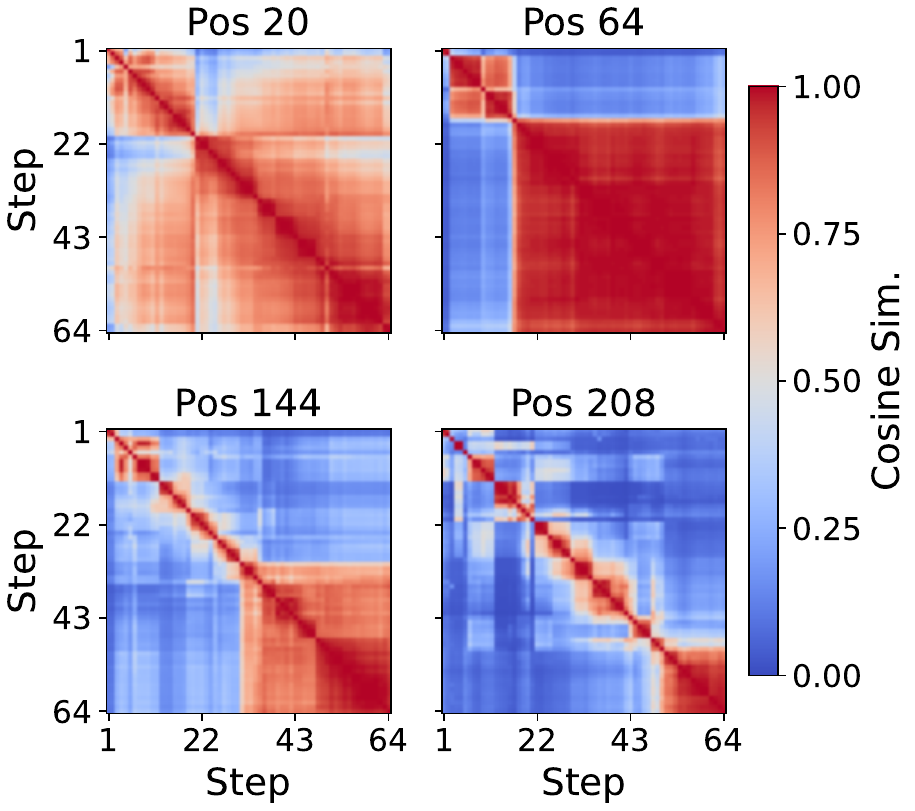}
        \vspace{-5mm}
        \caption{Cross-step cosine sim.}
        \label{fig:cross_step_heatmap}
    \end{subfigure}
    \begin{subfigure}[b]{\linewidth}
        \centering
        \includegraphics[width=\linewidth]{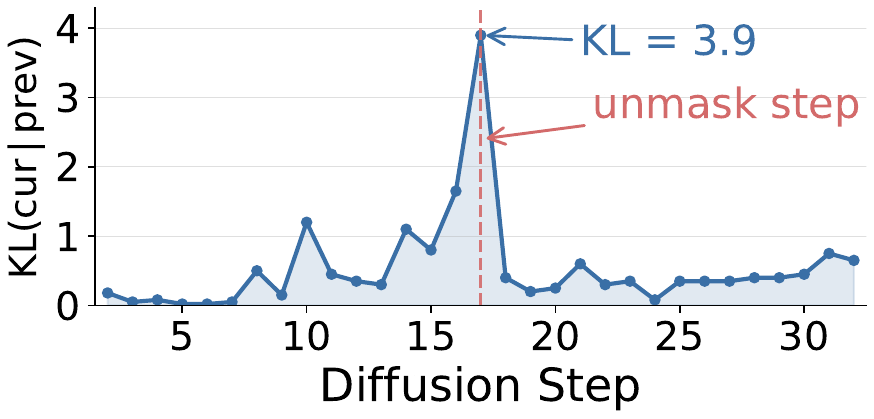}
        \vspace{-5mm}
        \caption{KL across steps.}
        \label{fig:cross_step_curve}
    \end{subfigure}
    \vspace{-5mm}
    \caption{Attention pattern dynamics across denoising steps.}
    \label{fig:cross_step}
    \vspace{-5mm}
\end{wrapfigure}
To verify the non-stationarity of attention patterns across denoising steps, we examine how the attention distribution of a query evolves along its denoising trajectory. Figure~\ref{fig:cross_step_heatmap} visualizes the cosine similarity of attention patterns between every pair of denoising steps for four query positions $\{20,64,144,208\}$ over $64$ steps. The heatmaps reveal a consistent block-wise structure across different query locations: similarity remains high \emph{within} contiguous step intervals but drops sharply at some specific steps. This indicates that attention patterns are \emph{locally} stationary but globally non-stationary. Figure~\ref{fig:cross_step_curve} further quantifies this behavior by plotting, for a representative query, the KL divergence between attention patterns at consecutive steps over $32$ steps. The curve stays small for most steps but exhibits a dominant spike (KL$\approx$3.9) at the step where the query token transitions from masked to unmasked, marking an abrupt shift in its attention pattern. These observations challenge the implicit assumption behind static mask reuse, namely that the top-$K$ important connections identified at early steps remain valid throughout all subsequent denoising steps. They instead motivate a strategy that reuses a sparse mask within each locally stationary interval for improved computational efficiency while refreshing it at the selected steps along the denoising trajectory.

Motivated by these dynamics, we propose \textbf{Step-dLLM}, a sparse attention framework for dLLMs that refreshes the mask at selected anchor steps and reuses it only within locally consistent regions, avoiding per-step reconstruction. We further observe pronounced head-wise heterogeneity: heads differ substantially in how concentrated their attention mass is, so a uniform per-head budget is suboptimal. Step-dLLM therefore combines anchor-step mask refresh with head-adaptive global budget allocation, realized through fused Triton kernels. In summary, our contributions are as follows:
\begin{itemize}
    \item We empirically identify two key properties of sparse attention in dLLMs: attention patterns are only \emph{locally} consistent across denoising steps, and different heads exhibit substantially different concentration behaviors.
    \item Building on these observations, we propose \textbf{Step-dLLM}, which periodically refreshes sparse masks, reuses them across nearby denoising steps, allocates sparsity budgets adaptively across heads, and realizes the mask-refresh and mask-reuse stages with fused Triton kernels.
    \item We evaluate Step-dLLM on LLaDA-1.5 and Dream-7B-Instruct, achieving a $2.78\times$ end-to-end speedup over dense FlashAttention on LLaDA-1.5 at 8K context with 256 denoising steps and $90\%$ sparsity, alongside improved accuracy--latency trade-offs over prior sparse-attention methods.
\end{itemize}

\section{Background and Related Work}
\label{sec:background}

\subsection{Preliminaries}
\label{sec:prelim}

Diffusion large language models generate text by iteratively unmasking a corrupted sequence over $S$ denoising steps. Let $\mathcal{V}$ be the vocabulary and $L$ the total sequence length, comprising $m$ prompt tokens and $L-m$ response tokens at positions $\mathcal{R}=\{m{+}1,\ldots,L\}$. Starting from $x^{(0)}$ in which all response positions are set to $[\mathrm{MASK}]$, each step $s$ applies a Transformer denoiser $f_\theta$ to produce per-token distributions $\pi_i^{(s)}=\mathrm{Softmax}\bigl(f_\theta(x^{(s-1)})_i\bigr)$ for $i\in\mathcal{R}$; a subset of currently masked positions is then unmasked by sampling from $\pi_i^{(s)}$ according to a decreasing mask-ratio schedule, while prompt tokens stay frozen. Because dLLMs predict \emph{all} masked tokens simultaneously, $f_\theta$ uses \textbf{bidirectional} self-attention over the full length-$L$ sequence at every step, incurring $\mathcal{O}(L^2)$ cost per step and precluding direct, exact reuse of the standard autoregressive KV cache. Reducing this per-step quadratic cost is the central goal of our work.

\subsection{Related Work}
\label{sec:related}

\paragraph{Diffusion language models.}
Discrete diffusion was first developed in D3PM~\citep{austin2021d3pm}, SEDD~\citep{lou2024sedd}, and MDLM~\citep{sahoo2024mdlm}, which together closed much of the gap with autoregressive models. LLaDA~\citep{nie2025llada} scaled the paradigm to 8B parameters with subsequent multimodal, mixture-of-experts, and preference-optimized variants~\citep{you2025lladav,llada-moe2025,zhu2025llada15}.

\paragraph{Efficient inference for dLLMs.}
Acceleration efforts follow two complementary directions: reducing the number of decoding passes via block-wise (semi-autoregressive) generation, which also enables KV caching across completed blocks~\citep{arriola2025block,wu2025fastdllm,d2f2025}; and reducing per-step attention cost via sparse attention. The closest prior work, SparseD~\citep{wang2025sparsed}, uses full attention during an initial warm-up fraction, constructs a sparse mask from these attention maps, and reuses it for the remaining steps. We show that the step-dependent drift of dLLMs attention makes such one-shot reuse suboptimal under high sparsity, and address this with anchor-step mask refresh and head-adaptive budget allocation; our method is orthogonal to block-wise generation.

\paragraph{Sparse attention.}
Classical sparse attention uses fixed patterns~\citep{child2019sparse,beltagy2020longformer,zaheer2020bigbird}, while recent dynamic methods construct block-sparse masks on the fly via online estimation, query-aware selection, or learned gating~\citep{jiang2024minference,tang2024quest,zhu2025tactic,gao2024seerattention,yuan2025nsa}, complemented by IO-aware kernels~\citep{dao2022flashattention,dao2023flashattention2,ye2025flashinfer,xu2025xattention}. All of these target \emph{causal} attention and rebuild the mask at every forward pass. Our work targets \emph{bidirectional} dLLMs attention and amortizes mask construction across denoising steps by exploiting the temporal stability unique to iterative denoising (Section~\ref{sec:mask_reuse}).

\section{Analysis and Motivations}
\label{sec:analysis}
In this section, we analyze attention pattern dynamics during dLLMs inference and present two observations that motivate the design of Step-dLLM.
\begin{figure}[ht]
    \centering
    \includegraphics[width=0.8\linewidth]{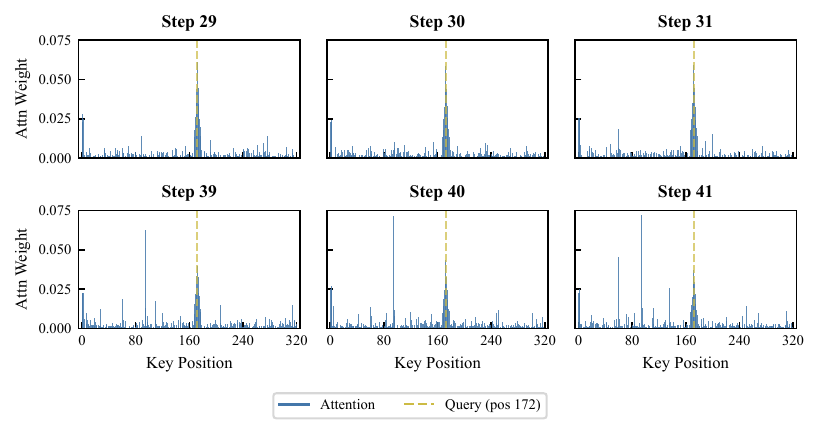}
    \caption{Attention pattern across different steps.}
    \label{fig:attention_pattern_across_steps}
\end{figure}
\subsection{Cross-step attention similarity}
Existing sparsity methods in dLLMs, such as SparseD \cite{wang2025sparsed}, are built on the observation that attention patterns exhibit noticeable temporal consistency across inference steps. Based on this observation, they estimate the sparse attention pattern once and reuse it in subsequent denoising steps, thereby avoiding repeated pattern construction overhead. While this assumption is partially valid, we find that it is insufficient to fully characterize attention dynamics in dLLMs. For a query position $i$ at denoising step $s$, we define its attention pattern as $\mathbf{a}_i^{(s)}=\mathrm{Softmax}\left(\frac{\mathbf{q}_i^{(s)}\mathbf{K}^{(s)\top}}{\sqrt{d}}\right)$, which represents the attention distribution of the query token over all key positions at step $s$.

Unlike autoregressive decoding, dLLMs inference proceeds through an iterative denoising process, in which the visible context and token certainty evolve continuously across steps. As more tokens become unmasked, the dependency structure of the sequence is gradually refined, and the attention pattern may be reorganized accordingly. In particular, some key positions that are not salient at early steps can become pivotal at later steps.  As illustrated in Figure~\ref{fig:attention_pattern_across_steps}, which compares two groups of nearby denoising steps, the attention patterns are highly consistent within Steps 29--31 and within Steps 39--41, suggesting local reusability across nearby steps. However, comparing these two groups reveals a clear pattern shift: a new pivotal token at key position 95 emerges in Steps 39--41 but is not prominent in Steps 29--31. This indicates that, although the sparse attention pattern can be reused locally, it is not fully stationary throughout denoising.

As further shown in Figure~\ref{fig:cross_step_curve}, the change between two consecutive attention patterns is measured by the KL divergence $D_{\mathrm{KL}}\!\left(\mathbf{a}_i^{(s)} \,\|\, \mathbf{a}_i^{(s-1)}\right)=\sum_{j=1}^{L}\mathbf{a}_{i,j}^{(s)}\log\frac{\mathbf{a}_{i,j}^{(s)}}{\mathbf{a}_{i,j}^{(s-1)}}$. Although many neighboring steps remain relatively similar, there exist several transition steps where the divergence increases sharply, suggesting abrupt changes in the underlying attention structure. This implies that directly reusing a sparse pattern estimated at early steps may lead to stale sparsity decisions: newly important attention links can be missed, while computation is still spent on entries that are no longer critical. These observations motivate us to refresh the sparse attention matrix at some selected anchor steps.

\begin{figure}[ht]
    \centering
    \begin{subfigure}[b]{0.35\columnwidth}
        \centering
        \includegraphics[width=\linewidth]{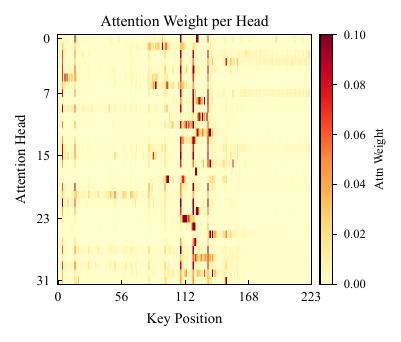}
        \caption{Attention patterns of different heads for the same query token.}
    \end{subfigure}
    \hspace{0.2\in}
    \begin{subfigure}[b]{0.35\columnwidth}
        \centering
        \includegraphics[width=\linewidth]{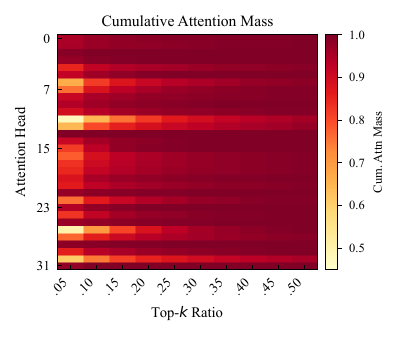}
        \caption{Cumulative attention mass across heads.}
        \label{fig:budget_cross_head}
    \end{subfigure}
    \caption{Head-wise heterogeneity in sparse attention.}
    \label{fig:attention_heterogeneity}
\end{figure}

\subsection{Head-wise Heterogeneity in Sparse Attention}
Head specialization has been observed in BERT~\citep{clark2019bert}, in-context learning~\citep{yin2025heads}, and head-wise memory architectures~\citep{kim2025midus}. These studies motivate treating heads individually; our analysis concerns their concentration behavior during dLLM denoising.
Recent sparse dLLM methods, such as SparseD~\cite{wang2025sparsed} and
Sparse-dLLM~\cite{song2026sparse}, reduce inference cost by selecting
important token interactions or evicting low-saliency cache entries at
the token level. These approaches implicitly treat the token as the
primary unit of sparsification, assuming that the importance of a token
is largely consistent across attention heads. We find that this
assumption does not hold in dLLMs.

As shown in Figure~\ref{fig:attention_heterogeneity}, attention heads
exhibit heterogeneity in two complementary aspects. First, for the same
query token, different heads attend to substantially different sets of
key positions: some focus sharply on a narrow region while others
distribute attention more broadly or emphasize entirely different tokens
(Figure~\ref{fig:attention_heterogeneity}(a)). Second, heads differ in
how concentrated their attention mass is: some are highly concentrated,
where a small fraction of top-scoring entries already retains most of
the attention mass, whereas others are much more dispersed and require
a substantially larger fraction to achieve the same coverage
(Figure~\ref{fig:attention_heterogeneity}(b)). This heterogeneity is
qualitatively different from the temporal patterns exploited by prior
work: SparseD pre-computes sparse patterns once and reuses them across
steps, while Sparse-dLLM leverages the stability of token saliency to
evict cache entries. These temporal mechanisms do not directly address the head-wise budget allocation studied here.

Together, these observations reveal that the key challenge in sparse
attention for dLLMs is not only \emph{which} token interactions to
preserve, but also \emph{how much} computation budget each head should
receive. This motivates us to allocate the sparse attention budget
globally across heads rather than enforcing a uniform per-head ratio,
allowing concentrated heads to be compressed more aggressively while
reserving more capacity for dispersed ones.

\section{Methodology}
\label{sec:method}
\begin{figure}[ht]
    \centering
    \includegraphics[width=0.9\linewidth]{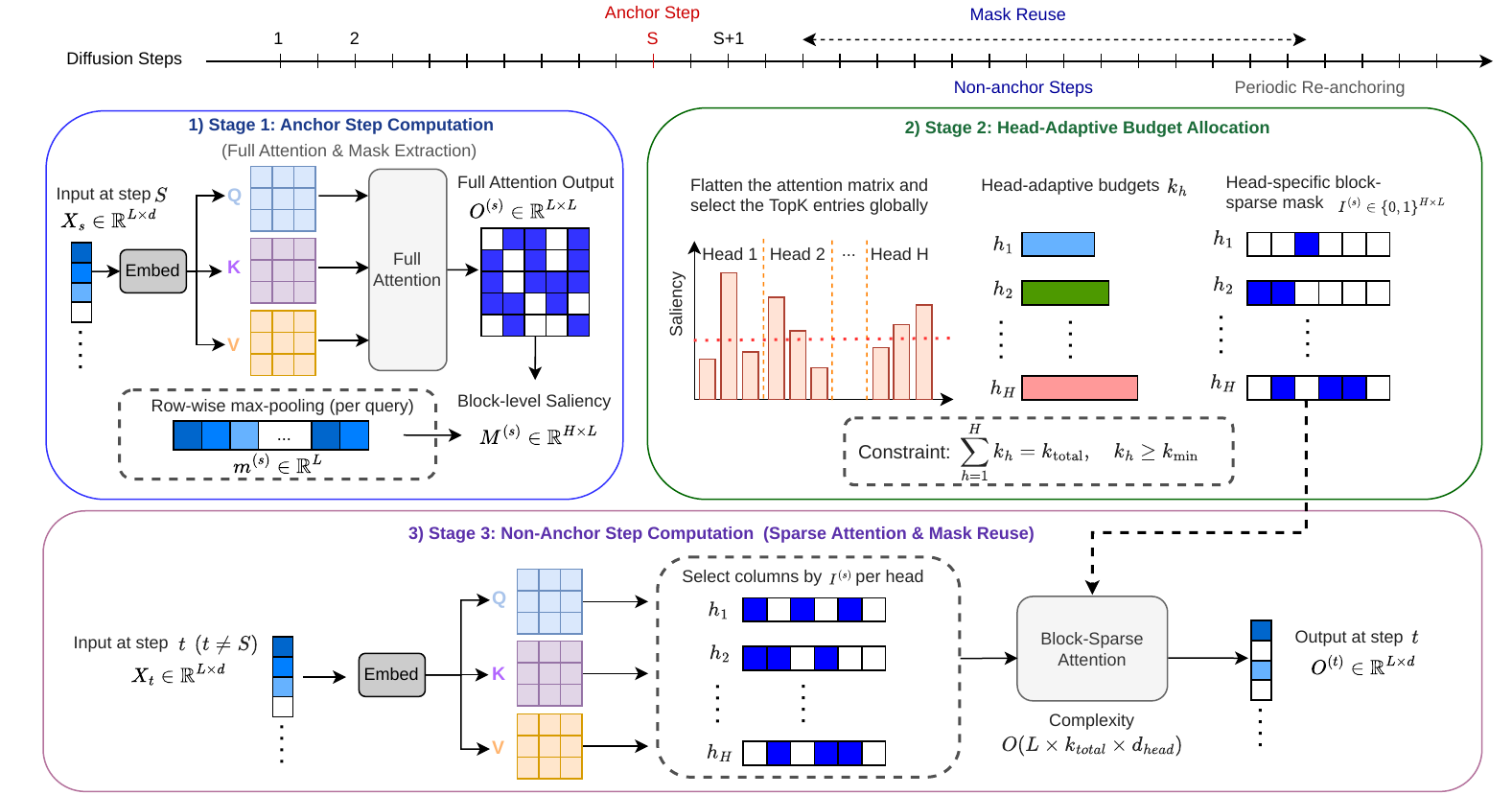}
    \caption{Overview of \textbf{Step-dLLM}. At an anchor step, the fused attention pass produces the dense output and block-level saliency (Stage 1). For each query block, saliency entries compete across all heads and key blocks for a shared top-$K$ budget; a minimum allocation yields head-specific sparse masks (Stage 2). The next $\Delta-1$ steps reuse only these masks, recomputing $Q,K,V$ and the attention weights from the current hidden states (Stage 3). A single query block per head is illustrated.}
    \label{fig:p}
\end{figure}

We present \textbf{Step-dLLM}, an efficient inference framework for dLLMs. It combines \emph{cross-step mask reuse} to amortize sparse-pattern identification with a \emph{head-adaptive budget} to allocate sparsity across attention heads. Both anchor and non-anchor steps are implemented with fused Triton kernels. Figure~\ref{fig:p} provides an overview.

\subsection{Cross-Step Sparse Mask Reuse}
\label{sec:mask_reuse}

As shown in Section~\ref{sec:analysis}, the attention patterns in dLLMs
are locally consistent across nearby denoising steps but are not globally
stationary: cumulative drift and occasional abrupt shifts make the one-shot
mask increasingly stale as denoising progresses. This motivates our strategy: \emph{reuse the sparse mask within several consecutive steps, but refresh it at some anchor steps} to prevent the
accumulated mismatch from degrading generation quality.

Formally, let $L$ denote the total sequence length, $b$ the block size, and $N_b = \lceil L/b \rceil$ the number of blocks. We partition the $S$ diffusion steps into consecutive groups of size $\Delta$. Within each group, the first step serves as the \emph{anchor} and the remaining $\Delta{-}1$ steps reuse the anchor's mask.

\paragraph{Anchor steps} compute full attention and extract block-level saliency $\mathbf{M}$ in the same fused pass. Budget allocation and top-$k$ selection then construct the mask $\mathcal{I}$ (Section~\ref{sec:mask_construction}):
\begin{equation}
    \mathbf{M},\ \mathbf{O}^{(s)} = \texttt{Modified-Flash-Attn}(\mathbf{Q}^{(s)}, \mathbf{K}^{(s)}, \mathbf{V}^{(s)},\ b).
\end{equation}

\paragraph{Non-anchor steps} freeze the cached mask $\mathcal{I}$ and restrict attention to the selected blocks. Note that only the sparsity pattern is inherited; $\mathbf{Q}^{(s)}$, $\mathbf{K}^{(s)}$, and $\mathbf{V}^{(s)}$ are freshly computed from the current step's hidden states, so the attention weights remain faithful to the evolving token representations:
\begin{equation}
    \mathbf{O}^{(s)} = \texttt{Block-Sparse-Attn}(\mathbf{Q}^{(s)}, \mathbf{K}^{(s)}, \mathbf{V}^{(s)},\ \mathcal{I}).
\end{equation}

The mask is refreshed every $\Delta$ steps. Let $k$ be the average number of retained key blocks per head and query block. Suppressing head count and feature dimension, the average attention cost is $\mathcal{O}(L^2/\Delta + Lkb)$, excluding mask-selection overhead. At fixed target sparsity, $k$ scales with $(1-\rho)L/b$, so attention remains quadratic in $L$ with a smaller coefficient. Algorithm~\ref{alg:step_aware_sparse} summarizes the procedure. The fixed schedule bounds refresh count without computing a drift signal at every step; Appendix~\ref{app:anchors} compares it with adaptive anchors.

\begin{algorithm}[ht]
\small
\caption{Step-dLLM: Step-Aware Block-Sparse Attention with Head-Adaptive Budget}
\label{alg:step_aware_sparse}
\begin{algorithmic}[1]
\Require Steps $S$, block size $b$, sparsity $\rho$, min budget $k_{\min}$, interval $\Delta$, heads $H$
\Ensure $\mathbf{O}^{(s)}$ for $s = 1, \ldots, S$
\State $N_b \gets \lceil L/b \rceil$;\quad $K \gets H \cdot \lfloor (1-\rho) N_b \rfloor$;\quad $\mathcal{I} \gets \varnothing$
    \Comment{\#blocks; per-query-block budget; cached mask}
\For{$s = 1$ \textbf{to} $S$}
    \State Compute $\mathbf{Q}^{(s)}, \mathbf{K}^{(s)}, \mathbf{V}^{(s)} \in \mathbb{R}^{H \times L \times d}$ from current hidden states
    \If{$\mathcal{I} = \varnothing$ \textbf{or} $s \equiv 1 \pmod{\Delta}$} \Comment{Anchor step}
        \State $\mathbf{M}, \mathbf{O}^{(s)} \gets \texttt{Modified-Flash-Attn}(\mathbf{Q}^{(s)}, \mathbf{K}^{(s)}, \mathbf{V}^{(s)}, b)$
            \Comment{$\mathbf{M} \in \mathbb{R}^{H \times N_b \times N_b}$}
        \For{$p = 1$ \textbf{to} $N_b$}
            \State $\mathcal{T}_p \gets \mathrm{Top\text{-}}K(\{\mathbf{M}^{(h)}_{pq}\}_{h,q})$;\quad
                   $c_h(p) \gets |\{(h',q)\in\mathcal{T}_p : h'=h\}|$
                \Comment{Global competition}
            \State $k_{h,p} \gets \max(c_h(p), k_{\min})$;\quad
                   $\mathcal{I}^{(h)}_p \gets \mathrm{Top\text{-}}k_{h,p}(\{\mathbf{M}^{(h)}_{pq}\}_q)$
                \Comment{Per-head budget \& sparse mask}
        \EndFor
    \Else \Comment{Non-anchor step}
        \State $\mathbf{O}^{(s)} \gets \texttt{Block-Sparse-Attn}(\mathbf{Q}^{(s)}, \mathbf{K}^{(s)}, \mathbf{V}^{(s)}, \mathcal{I})$
            \Comment{Reuse cached mask $\mathcal{I}$}
    \EndIf
\EndFor
\end{algorithmic}
\end{algorithm}

\subsection{Block-Sparse Mask Construction with Head-Adaptive Budget}
\label{sec:mask_construction}

We now describe how the sparsity mask $\mathcal{I}$ is constructed at each anchor step. The procedure involves two stages: computing block-level saliency scores and allocating a non-uniform budget across heads.

\paragraph{Block-level saliency scoring.}
For each attention head $h$, query block $p$, and key block $q$, we compute a coarse saliency score via block-wise max-pooling over the attention logits:
\begin{equation}
    \mathbf{M}^{(h)}_{pq} = \max_{(m,n) \in [b] \times [b]} \frac{(\mathbf{Q}^{(h)}_p)_m (\mathbf{K}^{(h)}_q)_n^\top}{\sqrt{d}},
\end{equation}
where $\mathbf{M}^{(h)}_{pq}$ captures the maximum attention affinity within each block pair for head $h$. This score is reduced from logits already computed in the full attention pass, avoiding a second read of $Q,K$ for saliency estimation; storing the block scores and selecting the mask still incur overhead.

\paragraph{Head-adaptive budget allocation.}
As shown in Section~\ref{sec:analysis}, different heads exhibit
distinct concentration behaviors: some heads focus
sharply on a few key blocks while others distribute attention
broadly. A uniform per-head budget cannot accommodate this
heterogeneity. We therefore determine each head's sparsity budget through
a global competition over saliency scores.

For each query block $p$, we flatten the saliency scores
$\{\mathbf{M}^{(h)}_{pq}\}$ across all heads into a single
vector of size $H \cdot N_b$ and select the top-$K$ entries,
where $K = H \cdot \lfloor (1-\rho) \cdot N_b \rfloor$ is the
budget derived from the target sparsity ratio $\rho$. Let
$c_h(p)$ denote the number of entries head $h$ wins at query
block $p$. The per-head budget for query $p$ is:
\begin{equation}
    k_{h,p} = \max\!\left(c_h(p),\; k_{\min}\right),
    \label{eq:head_budget}
\end{equation}
which ensures every head retains at least $k_{\min}$ blocks
for that query block. Each head then independently
constructs its sparse mask:
\begin{equation}
    \mathcal{I}^{(h)}_p = \mathrm{Top\text{-}k_{h,p}}\!\left(
    \{\mathbf{M}^{(h)}_{pq}\}_{q=1}^{N_b}\right), \quad
    \mathcal{I}^{(h)} =
    \{\mathcal{I}^{(h)}_p\}_{p=1}^{N_b}.
\end{equation}
The per-head allocation emerges from competition over saliency rather than an explicit dispersion metric. Because the scores are raw logits, their scales may differ across heads; this heuristic does not guarantee an allocation ordered by normalized attention concentration. Cross-head score calibration remains a limitation.

\section{Experiments}
\label{sec:experiments}

\subsection{Experimental Setup}
\textbf{Setup.} We apply Step-dLLM to two representative dLLMs, LLaDA-1.5~\cite{zhu2025llada15} and Dream-7B-Instruct~\cite{ye2025dream}, and evaluate on five benchmarks covering reasoning (GSM8K, MATH), code (HumanEval), and long-context retrieval (NIAH, RULER). We compare against the dense backbone, the AR-style sparse method StreamingLLM~\cite{xiao2023efficient}, and the dLLM-targeted sparse method SparseD~\cite{wang2025sparsed}. We report per-benchmark accuracy for quality, and kernel-level and end-to-end speedup over dense Flash Attention~\citep{dao2022flashattention} for efficiency.

\textbf{Implementation Details.}
All experiments are conducted on a single NVIDIA H100 GPU. For our proposed Step-dLLM, we largely follow
the hyperparameter conventions of SparseD~\cite{wang2025sparsed}. We set the attention block
size to $b = 16$ and the semi-autoregressive generation block length
to $B = 32$ for all datasets; the latter controls the ratio between
autoregressive steps and diffusion steps and is distinct from the
attention block size $b$. The re-anchor interval is set to
$\Delta = 0.2 \cdot S$, i.e., the sparse attention mask is refreshed
every $0.2 \cdot S$ denoising steps, corresponding to a refresh
interval ratio of $f = \Delta / S = 0.2$.
All reported experiments use $k_{\min}=1$, i.e., at least one key block per head and query block (not a fixed percentage). This floor can increase the realized keep rate above $1-\rho$; we report nominal sparsity and do not claim exact budget preservation after flooring. The global budget
$K_{\text{total}}$ is derived from the target sparsity ratio $\rho$
via $K_{\text{total}} = H \cdot \lfloor (1-\rho) \cdot N_b \rfloor$
(Section~\ref{sec:mask_construction}). For all benchmarks, we evaluate
under four sparsity ratios $\rho \in \{50\%, 80\%, 90\%, 95\%\}$.
SparseD uses the same $(b, B, \rho)$ configuration, with its
early-step full-attention skip ratio set to $5\%$.
For StreamingLLM, we fix the proportion
of sink tokens to $s_0 = 10\%$ of the total generation length and
derive the sliding window size $W$ accordingly. Model, dataset, and baseline descriptions are provided in Appendix~\ref{app:setup}.
\begin{table}[t]
\centering
\caption{Main results on LLaDA-1.5 and Dream-7B-Instruct (accuracy in \%). NIAH and RULER columns are averaged across their sub-tasks; per-sub-task scores are in Appendix~\ref{app:detailed-tables}. Within each sparsity block, \textbf{bold} marks the highest score per (backbone, metric). Step-dLLM rows are highlighted.}
\label{tab:main-bs16-llada}
\footnotesize
\setlength{\tabcolsep}{3pt}
\renewcommand{\arraystretch}{1.0}
\begin{tabular}{llccccccccccc}
\toprule
& & \multicolumn{5}{c}{LLaDA-1.5} & & \multicolumn{5}{c}{Dream-7B-Instruct} \\
\cmidrule(lr){3-7} \cmidrule(lr){9-13}
Sparsity & Method & GSM8K & Math & HEval & NIAH & RULER & & GSM8K & Math & HEval & NIAH & RULER \\
\midrule
\multicolumn{2}{l}{Dense} & 81.04 & 37.20 & 40.24 & 100.00 & 76.78 & & 77.41 & 45.60 & 59.12 & 95.06 & 81.33 \\
\midrule

50\% & StreamingLLM & 74.07 & 38.20 & 39.02 & 18.75 & 48.77 & & 38.13 &  8.80 & 42.70 & 10.16 & 41.77 \\
     & SparseD      & 80.59 & 37.40 & \textbf{42.68} & \textbf{100.00} & 74.86 & & 70.50 & 43.60 & 56.10 & 94.78 & \textbf{80.44} \\
\rowcolor{gray!15}
     & Step-dLLM    & \textbf{81.65} & \textbf{40.00} & \textbf{42.68} & \textbf{100.00} & \textbf{75.05} & & \textbf{71.80} & \textbf{44.00} & \textbf{58.54} & \textbf{94.81} & 79.75 \\
\midrule

80\% & StreamingLLM & 19.11 & 13.60 & 19.51 &  2.50 & 25.75 & & 19.56 & 10.40 & 25.00 &  1.84 &  9.60 \\
     & SparseD      & 53.68 & 29.20 & 36.59 & 88.79 & 71.75 & & 49.81 & 37.80 & 53.66 & 77.72 & 75.89 \\
\rowcolor{gray!15}
     & Step-dLLM    & \textbf{77.56} & \textbf{36.40} & \textbf{38.41} & \textbf{99.74} & \textbf{73.73} & & \textbf{68.46} & \textbf{38.60} & \textbf{54.27} & \textbf{85.59} & \textbf{76.15} \\
\midrule

90\% & StreamingLLM &  0.08 &  0.00 &  2.43 &  0.00 &  0.48 & &  0.00 &  0.00 &  2.44 &  0.00 &  0.00 \\
     & SparseD      & 29.64 & 20.00 & 30.49 & 89.63 & 65.82 & & 22.21 & 28.80 & 43.90 & \textbf{77.06} & 70.90 \\
\rowcolor{gray!15}
     & Step-dLLM    & \textbf{61.79} & \textbf{26.20} & \textbf{34.76} & \textbf{99.14} & \textbf{70.97} & & \textbf{32.68} & \textbf{32.00} & \textbf{50.00} & 76.12 & \textbf{72.35} \\
\midrule

95\%$^{\star}$ & SparseD      &  9.86 &  9.40 & 20.12 & 90.06 & 64.23 & &  4.40 & 14.80 & 31.10 & \textbf{74.60} & 64.47 \\
\rowcolor{gray!15}
     & Step-dLLM    & \textbf{18.80} & \textbf{16.60} & \textbf{23.17} & \textbf{96.84} & \textbf{66.78} & & \textbf{15.31} & \textbf{19.20} & \textbf{35.98} & 72.36 & \textbf{66.22} \\
\bottomrule
\end{tabular}
{\raggedright \scriptsize $^{\star}$StreamingLLM is omitted at $\rho = 95\%$ as all metrics collapse to zero on both backbones.\par}
\end{table}

\subsection{Accuracy}

Table~\ref{tab:main-bs16-llada} reports accuracy on LLaDA-1.5 and Dream-7B-Instruct under four sparsity ratios. For NIAH and RULER, we report the average over their sub-tasks; per-sub-task scores are in Appendix~\ref{app:detailed-tables}. Step-dLLM improves accuracy over SparseD in most settings, especially on reasoning and code tasks at high sparsity; the ordering on long-context tasks depends on the backbone and sub-task.

\paragraph{Reasoning and code generation.}
On GSM8K, MATH, and HumanEval, both sparse methods largely retain the dense baseline on LLaDA-1.5 at $\rho = 50\%$; Dream-7B-Instruct still exhibits a GSM8K gap to dense. The gap opens at higher sparsity. At $\rho = 90\%$ on GSM8K, Step-dLLM increases accuracy from $29.64\%$ to $61.79\%$ on LLaDA-1.5 and from $22.21\%$ to $32.68\%$ on Dream-7B-Instruct over SparseD. At $\rho = 95\%$, SparseD collapses to single-digit accuracy ($9.86\%$ and $4.40\%$). Step-dLLM retains $18.80\%$ and $15.31\%$. The same trend holds on MATH and HumanEval. This is because the one-shot mask in SparseD grows progressively stale as denoising proceeds. Periodic refresh re-aligns the sparse pattern with the evolving attention, which prevents the late-step collapse that SparseD suffers under tight budgets.

\paragraph{Long-context retrieval.}
On NIAH, Step-dLLM is essentially lossless up to $\rho = 90\%$ on LLaDA-1.5, increasing from $89.63\%$ for SparseD to $99.14\%$, and still retains $96.84\%$ at $\rho = 95\%$. On Dream-7B-Instruct, the NIAH gain holds at $\rho=80\%$, but SparseD is better at $90\%$ and $95\%$ (77.06 vs.\ 76.12 and 74.60 vs.\ 72.36). On RULER, Step-dLLM leads on LLaDA-1.5 at all evaluated sparsities and on Dream-7B-Instruct except at $50\%$ (e.g., from $65.82\%$ to $70.97\%$ on LLaDA-1.5 at $\rho = 90\%$). Appendix~\ref{app:detailed-tables} shows the gain concentrates on multi-needle and compositional sub-tasks. These sub-tasks depend on a few sparsely scattered key positions, so any mask staleness is amplified. As a result, periodic refresh together with head-adaptive budget allocation matters most exactly here.

In contrast, StreamingLLM collapses at high sparsity. On LLaDA-1.5 GSM8K, it drops from $74.07\%$ at $\rho{=}50\%$ to nearly zero at $\rho{=}90\%$, and we therefore omit its $\rho{=}95\%$ row. The reason is that StreamingLLM relies on causal-decoder priors (sink tokens plus a sliding window). These priors do not capture the bidirectional, step-evolving attention of dLLMs.

\subsection{Efficiency}
\begin{figure}[!ht]
    \centering
    \includegraphics[width=1\linewidth]{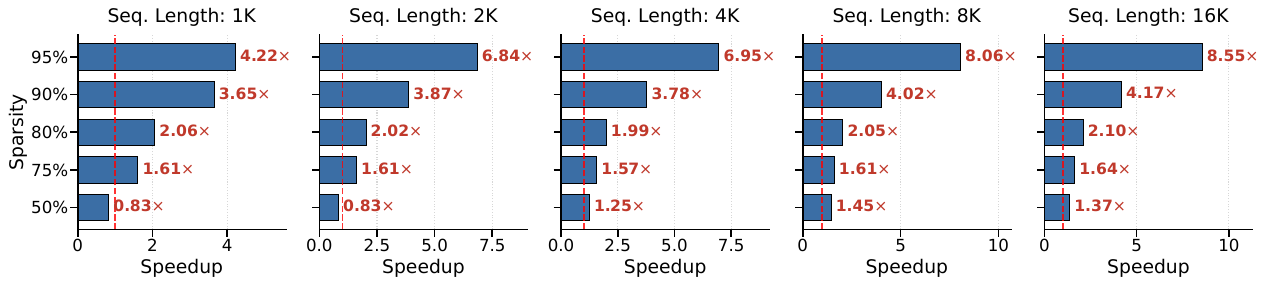}
    \caption{Kernel-level wall-clock speedup of the fused block-sparse attention kernel over the dense Flash Attention baseline.}
    \label{fig:kernel}
\end{figure}

\begin{figure}[!ht]
    \centering
    \includegraphics[width=1\linewidth]{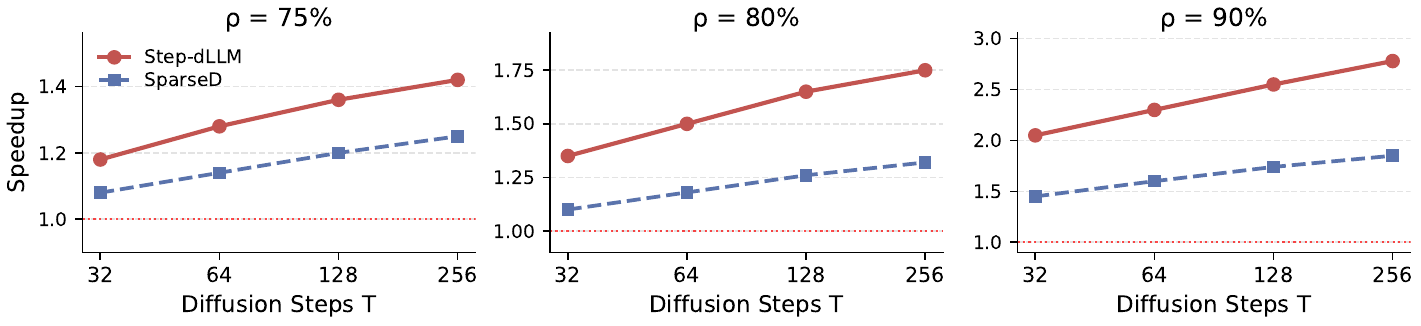}
    \caption{End-to-end speedup of LLaDA-1.5-8B over dense FlashAttention at 8K context.}
    \label{fig:e2e_latency}
\end{figure}

\paragraph{Kernel-level efficiency.}
We implement two fused Triton~\citep{tillet2019triton} kernels that save IO and memory in the spirit of Flash Attention~\citep{dao2022flashattention,dao2023flashattention2}. \texttt{Modified-Flash-Attn} extends Flash Attention~\citep{dao2022flashattention} to produce both $\mathbf{O}$ and the block-level saliency matrix $\mathbf{M}^{(h)}$ in a single pass. \texttt{Block-Sparse-Attn} consumes the cached indices $\mathcal{I}^{(h)}$ and loads only the $k_{h,p}$ selected key/value blocks per query block. The anchor pass avoids materializing a token-level $L\times L$ attention map, but stores a block-level saliency tensor with $HN_b^2$ entries; at fixed $b$, this tensor still scales quadratically with $L$. As shown in Figure~\ref{fig:kernel}, the block-sparse kernel achieves $8.55\times$ over dense Flash Attention at $16$K under $\rho = 95\%$ and around $2\times$ at $\rho = 80\%$ across all tested lengths. The reason is that the fused kernels turn algorithmic block-sparsity into real SRAM and HBM savings. At $50\%$ sparsity the kernel instead reaches only $0.83\times$ at 1K and 2K: irregular K/V gathers and variable per-tile work can outweigh the reduced FLOPs. Kernel timings do not include the amortized anchor and mask-selection costs.
\paragraph{End-to-end efficiency.}
We further measure end-to-end latency of LLaDA-1.5-8B at $8$K context across $S \in \{32, 64, 128, 256\}$. As shown in Figure~\ref{fig:e2e_latency}, Step-dLLM beats SparseD at every $(\rho, S)$ pair. At $S{=}256$, it reaches $1.42\times$, $1.75\times$, and $2.78\times$ over FlashAttention at $\rho = 75\%, 80\%, 90\%$, and the speedup grows with both $S$ and $\rho$. The reason is that our fused IO-aware kernel skips unselected key/value blocks rather than masking them after a dense load. These measurements include dense anchor steps. Their speedups depend on context length, sparsity, refresh interval, and the fraction of runtime spent outside attention; kernel-level gains do not imply proportional generation speedups.
\paragraph{Joint accuracy--latency trade-off.}
Appendix~\ref{app:tradeoff} reports the complete GSM8K sweep at approximately 1K sequence length, including $\rho=50\%$. Dense reaches 81.04\% at 50.39 ms/step. Step-dLLM at $\rho=70\%$ reaches 80.74\% at 48.50 ms/step, compared with SparseD's 80.59\% at 59.94 ms/step at $\rho=50\%$: a $1.24\times$ latency advantage at similar accuracy. At $\rho=50\%$, Step-dLLM takes 51.23 ms/step, slightly slower than dense. Thus the practical gain over dense here is modest ($1.04\times$ at $70\%$), distinct from the larger 8K-context speedups. Small accuracy gains over dense may reflect evaluation variability or pruning of weak connections; they do not establish a denoising mechanism or statistical significance.

\paragraph{Implementation backends and contribution.}
Both sparse paths use fused Triton kernels: SparseD uses PyTorch \texttt{flex\_attention}, compiled by TorchInductor, whereas Step-dLLM adapts SeerAttention's block-sparse kernel to heterogeneous per-head budgets. SparseD's mask construction operates on materialized attention maps and computes the dense output separately; our anchor pass combines dense output and block-saliency extraction. The accuracy contribution is the refresh and budget algorithm; efficiency reflects its co-design with the kernels, rather than the use of Triton alone. This is a comparison of complete implementations, not an isolation of algorithmic speed under an identical backend.

\subsection{Ablation Studies}

We ablate the two design choices in Step-dLLM, periodic mask refresh and head-adaptive budget allocation, and study sensitivity to the block size. Refresh and block-size ablations are conducted on LLaDA-1.5. Tables~\ref{tab:ablation-freq} and~\ref{tab:ablation-blocksize} include $\rho=50\%$; Appendix~\ref{app:pooling} compares saliency pooling; Appendix~\ref{app:cache} evaluates composition with Fast-dLLM dual caching.
\begin{table}[!ht]
\centering
\caption{Ablation on mask refresh interval. $f=\Delta/S$; lower $f$ means more frequent refresh. For $f<1$, SparseD denotes its refresh-augmented, uniform-budget variant. Accuracy in \% on LLaDA-1.5, including $50\%$ sparsity. G8K, VT denote GSM8K and RULER-VT; bold marks the better method within each configuration (ties included).}
\label{tab:ablation-freq}
\footnotesize
\setlength{\tabcolsep}{2.5pt}
\renewcommand{\arraystretch}{1.0}
\begin{tabular}{llrrrrrrrrrrrr}
\toprule
\multirow{2}{*}{Interval} & \multirow{2}{*}{Method} & \multicolumn{3}{c}{Sparsity $50\%$} & \multicolumn{3}{c}{Sparsity $80\%$} & \multicolumn{3}{c}{Sparsity $90\%$} & \multicolumn{3}{c}{Sparsity $95\%$} \\
\cmidrule(lr){3-5}\cmidrule(lr){6-8}\cmidrule(lr){9-11}\cmidrule(lr){12-14}
& & G8K & MATH & VT & G8K & MATH & VT & G8K & MATH & VT & G8K & MATH & VT \\
\midrule
$f{=}1.0$ & SparseD & \textbf{80.59} & 37.40 & 99.88 & 53.68 & 29.20 & 94.32 & 29.64 & \textbf{23.00} & 93.64 & 9.48 & 12.00 & 86.16 \\
\rowcolor{gray!15}
 & Step-dLLM & 77.48 & \textbf{40.20} & \textbf{100.00} & \textbf{57.77} & \textbf{31.00} & \textbf{98.52} & \textbf{35.18} & 21.20 & \textbf{94.00} & \textbf{16.83} & \textbf{15.40} & \textbf{92.32} \\
\addlinespace[2pt]
$f{=}0.5$ & SparseD & \textbf{81.50} & \textbf{40.60} & \textbf{100.00} & 64.37 & 31.20 & 95.12 & 29.72 & 22.20 & \textbf{97.04} & 6.29 & 13.80 & \textbf{95.40} \\
\rowcolor{gray!15}
 & Step-dLLM & 81.27 & \textbf{40.60} & \textbf{100.00} & \textbf{68.23} & \textbf{32.00} & \textbf{96.64} & \textbf{40.64} & \textbf{22.60} & 96.56 & \textbf{16.98} & \textbf{16.00} & 95.36 \\
\addlinespace[2pt]
$f{=}0.3$ & SparseD & 82.03 & \textbf{40.80} & \textbf{100.00} & 73.09 & 32.40 & 95.23 & 34.19 & 23.20 & \textbf{97.96} & 5.53 & 14.20 & \textbf{97.04} \\
\rowcolor{gray!15}
 & Step-dLLM & \textbf{82.34} & \textbf{40.80} & \textbf{100.00} & \textbf{75.01} & \textbf{33.60} & \textbf{97.28} & \textbf{52.24} & \textbf{23.60} & 97.04 & \textbf{17.43} & \textbf{15.40} & 96.36 \\
\addlinespace[2pt]
$f{=}0.2$ & SparseD & 80.97 & \textbf{40.60} & \textbf{100.00} & 76.80 & 34.00 & 95.80 & 43.90 & 24.80 & \textbf{97.96} & 6.29 & 13.60 & 96.64 \\
\rowcolor{gray!15}
 & Step-dLLM & \textbf{81.65} & 40.00 & 99.60 & \textbf{77.56} & \textbf{36.40} & \textbf{99.88} & \textbf{61.79} & \textbf{26.20} & 97.50 & \textbf{18.80} & \textbf{16.60} & \textbf{97.20} \\
\bottomrule
\end{tabular}
\end{table}

\paragraph{Refresh frequency.}
Table~\ref{tab:ablation-freq} reports performance under varying refresh intervals $f$. On GSM8K at $\rho = 90\%$, SparseD increases from $29.64\%$ to $43.90\%$ between $f = 1.0$ and $f = 0.2$. Step-dLLM increases from $35.18\%$ to $61.79\%$ over the same interval. Refresh re-aligns the mask with the current attention pattern. At $\rho=50\%$, neither method consistently dominates: GSM8K is 80.59 vs.\ 77.48 at $f=1.0$, but 80.97 vs.\ 81.65 at $f=0.2$ (SparseD vs.\ Step-dLLM). Thus the advantage is most pronounced under tight budgets.

\paragraph{Head-adaptive budget.}
Head-adaptive allocation contributes independently of refresh. At $f=1.0$, neither method refreshes its mask; the head-adaptive versus uniform allocation is the only algorithmic difference in this ablation. On GSM8K at $\rho = 90\%$ it increases from $29.64\%$ to $35.18\%$, and on RULER-VT at $\rho = 95\%$ from $86.16\%$ to $92.32\%$. The reason is that a uniform per-head budget over-allocates blocks to heads with concentrated mass and under-allocates blocks to dispersed heads. Redistributing toward dispersed heads yields a better mask before any refresh is introduced.
\begin{table}[!ht]
\centering
\caption{Ablation on attention block size $b$. Accuracy in \% on LLaDA-1.5, including $50\%$ sparsity. G8K, HE denote GSM8K and HumanEval; bold marks the better method within each configuration (ties included).}
\label{tab:ablation-blocksize}
\footnotesize
\setlength{\tabcolsep}{2.5pt}
\renewcommand{\arraystretch}{1.0}
\begin{tabular}{llrrrrrrrrrrrr}
\toprule
\multirow{2}{*}{Block} & \multirow{2}{*}{Method} & \multicolumn{3}{c}{Sparsity $50\%$} & \multicolumn{3}{c}{Sparsity $80\%$} & \multicolumn{3}{c}{Sparsity $90\%$} & \multicolumn{3}{c}{Sparsity $95\%$} \\
\cmidrule(lr){3-5}\cmidrule(lr){6-8}\cmidrule(lr){9-11}\cmidrule(lr){12-14}
& & G8K & MATH & HE & G8K & MATH & HE & G8K & MATH & HE & G8K & MATH & HE \\
\midrule
$b{=}16$ & SparseD & 80.59 & 37.40 & \textbf{42.68} & 53.68 & 29.20 & 36.59 & 29.64 & 20.00 & 30.49 & 9.86 & 9.40 & 20.12 \\
\rowcolor{gray!15}
 & Step-dLLM & \textbf{81.65} & \textbf{40.00} & \textbf{42.68} & \textbf{77.56} & \textbf{36.40} & \textbf{38.41} & \textbf{61.79} & \textbf{26.20} & \textbf{34.76} & \textbf{18.80} & \textbf{16.60} & \textbf{23.17} \\
\addlinespace[2pt]
$b{=}32$ & SparseD & \textbf{81.73} & 34.80 & \textbf{38.41} & 50.87 & \textbf{31.60} & 37.80 & 17.51 & 21.60 & 28.66 & 8.64 & 10.60 & 12.80 \\
\rowcolor{gray!15}
 & Step-dLLM & 81.58 & \textbf{41.00} & 37.80 & \textbf{63.23} & 31.20 & \textbf{39.12} & \textbf{39.50} & \textbf{26.00} & \textbf{34.15} & \textbf{19.18} & \textbf{16.20} & \textbf{23.78} \\
\bottomrule
\end{tabular}
\end{table}

\paragraph{Block size.}
Table~\ref{tab:ablation-blocksize} compares $b = 16$ and $b = 32$. Step-dLLM improves GSM8K at both block sizes; the absolute margin depends on sparsity. On GSM8K at $\rho = 90\%$, Step-dLLM increases accuracy from $29.64\%$ to $61.79\%$ at $b = 16$ and from $17.51\%$ to $39.50\%$ at $b = 32$ over SparseD. At $50\%$, gains are mixed: Step-dLLM improves MATH at both block sizes, while SparseD slightly leads GSM8K and HumanEval at $b=32$. Larger blocks remove more tokens per selection, motivating a joint choice of block size and sparsity.

\section{Conclusion and Limitations}
\label{sec:conclusion}
We presented \textbf{Step-dLLM}, a training-free sparse attention framework for diffusion LLMs. Motivated by local cross-step consistency and head-wise concentration heterogeneity, Step-dLLM refreshes a block-sparse mask at periodic anchor steps and uses a head-adaptive global budget, backed by a fused IO-aware Triton kernel. Experiments on LLaDA-1.5 and Dream-7B-Instruct demonstrate improved accuracy--latency trade-offs over prior sparse-attention methods. On LLaDA-1.5 at 8K context with 256 denoising steps and $90\%$ sparsity, Step-dLLM achieves a $2.78\times$ end-to-end speedup over dense FlashAttention.

\paragraph{Limitations.} Step-dLLM uses a heuristic anchor schedule rather than drift-triggered adaptive anchors, and is evaluated only on text-only dLLMs. The joint accuracy--latency sweep covers one model, task, and short-context setting; wider operating-point validation is needed. Raw-logit saliency may be affected by head-wise scale differences, and the minimum-head floor can relax nominal sparsity. Adaptive triggers and extension to multimodal diffusion models remain future work.

\newpage
\bibliographystyle{plainnat}
\bibliography{example_paper}

\clearpage
\appendix
\section{Detailed Per-Sub-Task Results}
\label{app:detailed-tables}

This appendix lists the per-sub-task scores for the long-context benchmarks summarized in the main paper (Table~\ref{tab:main-bs16-llada}). For each backbone, we separate the eight NIAH sub-tasks from the five RULER sub-tasks so that each table can be read at a normal font size. The ``NIAH (avg)'' and ``RULER (avg)'' columns in the main paper are obtained by averaging the values listed here, weighting all sub-tasks equally.

\paragraph{NIAH sub-tasks.} The Needle-in-a-Haystack benchmark probes how reliably a model can locate a small piece of information that has been planted inside a long context. We evaluate eight variants. \textit{Single-needle} tasks (\textbf{S\_1, S\_2, S\_3}) ask the model to retrieve a single answer planted at one position; the three variants differ in needle length and document complexity. \textit{Multi-key} tasks (\textbf{MK\_1, MK\_2, MK\_3}) plant several distractor keys alongside the target so the model must disambiguate among multiple candidates. \textit{Multi-query} (\textbf{MQ}) requires the model to answer several queries against the same context in one pass, while \textit{multi-value} (\textbf{MV}) requires extracting several values associated with the same key. Together these settings cover both retrieval accuracy and the ability to track multiple pieces of evidence simultaneously.

\paragraph{RULER sub-tasks.} RULER stresses long-context reasoning beyond simple retrieval. \textbf{VT} (variable tracking) asks the model to follow chains of variable assignments scattered across the context. \textbf{CWE} and \textbf{FWE} (common- and frequent-word extraction) probe whether the model can identify globally distinguished tokens by aggregating statistics over the entire input. \textbf{QS} and \textbf{QH} are single-document and multi-document QA, with the latter requiring the model to combine evidence from several passages. These sub-tasks therefore stress different aspects of long-context attention than NIAH: tracking, aggregation, and multi-hop composition.

\paragraph{Reading the tables.} Within each sparsity block, the highest score per sub-task is marked in bold; Step-dLLM rows are shaded. The Dense row is provided as a non-sparse reference and is never bolded. Tables~\ref{tab:main-bs16-llada-detailed-niah} and~\ref{tab:main-bs16-llada-detailed-ruler} cover LLaDA-1.5; Tables~\ref{tab:main-bs16-dream-detailed-niah} and~\ref{tab:main-bs16-dream-detailed-ruler} cover Dream-7B-Instruct. A few patterns are worth noting beyond the main paper. First, the gap between Step-dLLM and SparseD is concentrated on the harder multi-needle and multi-value tasks (S\_3, MK\_3, MV); on the easier single-needle tasks both methods stay near the dense baseline up to high sparsity. Second, on RULER the relative ordering between methods varies more across sub-tasks: Step-dLLM dominates on tracking-heavy tasks (VT, CWE) where the salient key positions shift across denoising steps, while SparseD remains competitive on QA-style tasks where the relevant context tends to be more localized.

\begin{table}[!ht]
\centering
\caption{Per-sub-task NIAH results on LLaDA-1.5 (accuracy in \%). The eight sub-tasks are single-needle (S\_1, S\_2, S\_3), multi-key (MK\_1, MK\_2, MK\_3), multi-query (MQ), and multi-value (MV). Within each sparsity block, \textbf{bold} marks the highest score per sub-task. Step-dLLM rows are highlighted.}
\label{tab:main-bs16-llada-detailed-niah}
\small
\setlength{\tabcolsep}{4pt}
\renewcommand{\arraystretch}{1.0}

\begin{tabular}{llcccccccc}
\toprule
Sparsity & Method & S\_1 & S\_2 & S\_3 & MK\_1 & MK\_2 & MK\_3 & MQ & MV \\
\midrule
\multicolumn{2}{l}{Dense} & 100.00 & 100.00 & 100.00 & 100.00 & 100.00 & 100.00 & 100.00 & 100.00 \\
\midrule

50\% & StreamingLLM & 20.80 & 20.20 & 18.40 & 21.20 & 20.40 & 15.40 & 19.20 & 14.40 \\
     & SparseD      & \textbf{100.00} & \textbf{100.00} & \textbf{100.00} & \textbf{100.00} & \textbf{100.00} & \textbf{100.00} & \textbf{100.00} & \textbf{100.00} \\
\rowcolor{gray!15}
     & Step-dLLM    & \textbf{100.00} & \textbf{100.00} & \textbf{100.00} & \textbf{100.00} & \textbf{100.00} & \textbf{100.00} & \textbf{100.00} & \textbf{100.00} \\
\midrule

80\% & StreamingLLM & 5.00 & 2.40 & 1.40 & 3.40 & 3.60 & 0.00 & 2.10 & 2.10 \\
     & SparseD      & \textbf{100.00} & \textbf{100.00} & 71.20 & \textbf{100.00} & \textbf{100.00} & 74.20 & 85.85 & 79.10 \\
\rowcolor{gray!15}
     & Step-dLLM    & \textbf{100.00} & \textbf{100.00} & \textbf{100.00} & \textbf{100.00} & 99.80 & \textbf{99.60} & \textbf{99.90} & \textbf{98.60} \\
\midrule

90\% & StreamingLLM & 0.00 & 0.00 & 0.00 & 0.00 & 0.00 & 0.00 & 0.00 & 0.00 \\
     & SparseD      & \textbf{100.00} & \textbf{100.00} & 76.60 & \textbf{100.00} & 99.80 & 77.20 & 82.85 & 80.60 \\
\rowcolor{gray!15}
     & Step-dLLM    & \textbf{100.00} & \textbf{100.00} & \textbf{100.00} & \textbf{100.00} & \textbf{100.00} & \textbf{97.60} & \textbf{99.65} & \textbf{95.90} \\
\midrule

95\% & SparseD      & \textbf{100.00} & \textbf{100.00} & 79.40 & \textbf{100.00} & \textbf{100.00} & 78.20 & 81.80 & 81.10 \\
\rowcolor{gray!15}
     & Step-dLLM    & \textbf{100.00} & \textbf{100.00} & \textbf{98.80} & \textbf{100.00} & \textbf{100.00} & \textbf{88.40} & \textbf{95.25} & \textbf{92.30} \\

\bottomrule
\end{tabular}
\end{table}

\begin{table}[!ht]
\centering
\caption{Per-sub-task RULER results on LLaDA-1.5 (accuracy in \%). The five sub-tasks are variable tracking (VT), common-word extraction (CWE), frequent-word extraction (FWE), single QA (QS), and hard QA (QH). Within each sparsity block, \textbf{bold} marks the highest score per sub-task. Step-dLLM rows are highlighted.}
\label{tab:main-bs16-llada-detailed-ruler}
\small
\setlength{\tabcolsep}{6pt}
\renewcommand{\arraystretch}{1.0}

\begin{tabular}{llccccc}
\toprule
Sparsity & Method & VT & CWE & FWE & QS & QH \\
\midrule
\multicolumn{2}{l}{Dense} & 100.00 & 52.40 & 64.20 & 91.30 & 76.00 \\
\midrule

50\% & StreamingLLM & 30.60 & 13.40 & \textbf{73.80} & 81.65 & 44.40 \\
     & SparseD      & \textbf{99.88} & 52.26 & 61.40 & \textbf{83.17} & 77.60 \\
\rowcolor{gray!15}
     & Step-dLLM    & 99.60 & \textbf{53.16} & 60.87 & 83.03 & \textbf{78.60} \\
\midrule

80\% & StreamingLLM & 6.56 & 3.02 & \textbf{76.27} & 20.48 & 22.40 \\
     & SparseD      & 94.28 & \textbf{57.80} & 45.40 & 83.48 & 77.80 \\
\rowcolor{gray!15}
     & Step-dLLM    & \textbf{98.52} & 56.46 & 51.73 & \textbf{83.73} & \textbf{78.20} \\
\midrule

90\% & StreamingLLM & 0.00 & 2.12 & 0.00 & 0.27 & 0.00 \\
     & SparseD      & 93.32 & 34.56 & 42.07 & 83.13 & 76.00 \\
\rowcolor{gray!15}
     & Step-dLLM    & \textbf{94.00} & \textbf{54.16} & \textbf{46.80} & \textbf{83.48} & \textbf{76.40} \\
\midrule

95\% & SparseD      & 86.64 & 39.92 & 39.00 & \textbf{81.97} & \textbf{73.60} \\
\rowcolor{gray!15}
     & Step-dLLM    & \textbf{92.32} & \textbf{44.76} & \textbf{44.07} & 81.37 & 71.40 \\

\bottomrule
\end{tabular}
\end{table}

\begin{table}[!ht]
\centering
\caption{Per-sub-task NIAH results on Dream-7B-Instruct (accuracy in \%). Within each sparsity block, \textbf{bold} marks the highest score per sub-task. Step-dLLM rows are highlighted.}
\label{tab:main-bs16-dream-detailed-niah}
\small
\setlength{\tabcolsep}{4pt}
\renewcommand{\arraystretch}{1.0}

\begin{tabular}{llcccccccc}
\toprule
Sparsity & Method & S\_1 & S\_2 & S\_3 & MK\_1 & MK\_2 & MK\_3 & MQ & MV \\
\midrule
\multicolumn{2}{l}{Dense} & 100.00 & 100.00 & 91.00 & 98.00 & 99.80 & 77.60 & 97.70 & 96.40 \\
\midrule

50\% & StreamingLLM & 10.00 & 11.00 & 9.00 & 11.00 & 8.80 & 12.40 & 8.75 & 10.30 \\
     & SparseD      & \textbf{100.00} & \textbf{100.00} & 90.20 & \textbf{98.00} & 99.40 & \textbf{75.80} & \textbf{97.45} & 97.40 \\
\rowcolor{gray!15}
     & Step-dLLM    & \textbf{100.00} & \textbf{100.00} & \textbf{90.50} & 97.80 & \textbf{99.80} & 75.60 & 97.25 & \textbf{97.50} \\
\midrule

80\% & StreamingLLM & 1.80 & 2.80 & 1.40 & 3.20 & 1.20 & 0.00 & 1.95 & 2.40 \\
     & SparseD      & \textbf{100.00} & \textbf{99.80} & 47.20 & \textbf{98.80} & \textbf{99.80} & 46.80 & 60.05 & 69.30 \\
\rowcolor{gray!15}
     & Step-dLLM    & \textbf{100.00} & \textbf{99.80} & \textbf{69.00} & 98.20 & \textbf{99.80} & \textbf{65.40} & \textbf{74.10} & \textbf{78.40} \\
\midrule

90\% & StreamingLLM & 0.00 & 0.00 & 0.00 & 0.00 & 0.00 & 0.00 & 0.00 & 0.00 \\
     & SparseD      & \textbf{100.00} & \textbf{100.00} & 44.60 & \textbf{98.80} & \textbf{99.80} & \textbf{44.80} & \textbf{59.45} & \textbf{69.00} \\
\rowcolor{gray!15}
     & Step-dLLM    & \textbf{100.00} & 99.80 & \textbf{47.00} & 97.20 & \textbf{99.80} & 39.80 & 57.30 & 68.05 \\
\midrule

95\% & SparseD      & \textbf{100.00} & \textbf{100.00} & \textbf{40.80} & \textbf{98.40} & 99.40 & \textbf{37.40} & \textbf{53.95} & \textbf{66.85} \\
\rowcolor{gray!15}
     & Step-dLLM    & \textbf{100.00} & 99.60 & 37.20 & 95.60 & \textbf{99.60} & 33.60 & 52.30 & 60.95 \\

\bottomrule
\end{tabular}
\end{table}

\begin{table}[!ht]
\centering
\caption{Per-sub-task RULER results on Dream-7B-Instruct (accuracy in \%). Within each sparsity block, \textbf{bold} marks the highest score per sub-task. Step-dLLM rows are highlighted.}
\label{tab:main-bs16-dream-detailed-ruler}
\small
\setlength{\tabcolsep}{6pt}
\renewcommand{\arraystretch}{1.0}

\begin{tabular}{llccccc}
\toprule
Sparsity & Method & VT & CWE & FWE & QS & QH \\
\midrule
\multicolumn{2}{l}{Dense} & 96.32 & 87.28 & 79.47 & 79.57 & 64.00 \\
\midrule

50\% & StreamingLLM & 13.40 & 43.34 & 37.73 & \textbf{81.40} & 33.00 \\
     & SparseD      & 96.28 & \textbf{85.16} & \textbf{75.80} & 80.37 & \textbf{64.60} \\
\rowcolor{gray!15}
     & Step-dLLM    & \textbf{96.40} & 83.24 & 75.53 & 79.80 & 63.80 \\
\midrule

80\% & StreamingLLM & 0.36 & 4.54 & 4.33 & 19.18 & 19.60 \\
     & SparseD      & 95.44 & 71.76 & \textbf{71.77} & \textbf{79.50} & 61.00 \\
\rowcolor{gray!15}
     & Step-dLLM    & \textbf{96.04} & \textbf{73.24} & 71.13 & 78.53 & \textbf{61.80} \\
\midrule

90\% & StreamingLLM & 0.00 & 0.00 & 0.00 & 0.00 & 0.00 \\
     & SparseD      & 92.20 & 54.72 & 67.53 & 78.45 & \textbf{61.60} \\
\rowcolor{gray!15}
     & Step-dLLM    & \textbf{94.68} & \textbf{59.58} & \textbf{68.26} & \textbf{79.03} & 60.20 \\
\midrule

95\% & SparseD      & 78.64 & 43.84 & 62.26 & 77.60 & 60.00 \\
\rowcolor{gray!15}
     & Step-dLLM    & \textbf{82.08} & \textbf{46.90} & \textbf{62.87} & \textbf{77.83} & \textbf{61.40} \\

\bottomrule
\end{tabular}
\end{table}

\clearpage
\section{Additional Accuracy--Efficiency Results}
\label{app:tradeoff}
We further evaluate the accuracy--latency trade-off. This sweep uses GSM8K, LLaDA-1.5, approximately 1K sequence length, and $f=0.2$. Table~\ref{tab:tradeoff} pairs accuracy with end-to-end inference time per denoising step. These are model-level ms/step measurements, not attention-kernel timings or full-generation durations. The 8K-context generation speedups in Figure~\ref{fig:e2e_latency} use a separate setup and should not be substituted into this table.

\begin{table}[ht]
\centering
\caption{Joint accuracy--latency sweep on GSM8K (LLaDA-1.5, approximately 1K sequence length). Accuracy is in percent; lower ms/step is better. Dense FlashAttention reaches 81.04\% at 50.39 ms/step.}
\label{tab:tradeoff}
\small
\setlength{\tabcolsep}{10pt}
\begin{tabular}{r r r r r}
\toprule
& \multicolumn{2}{c}{Step-dLLM} & \multicolumn{2}{c}{SparseD}\\
\cmidrule(lr){2-3}\cmidrule(lr){4-5}
Sparsity & Accuracy & ms/step & Accuracy & ms/step\\
\midrule
50\% & 81.65 & 51.23 & 80.59 & 59.94\\
60\% & 82.34 & 50.21 & 74.68 & 59.53\\
70\% & 80.74 & 48.50 & 63.15 & 59.15\\
75\% & 78.92 & 48.19 & 62.02 & 58.53\\
80\% & 77.56 & 47.36 & 53.68 & 58.06\\
85\% & 74.91 & 47.57 & 42.76 & 57.69\\
90\% & 61.79 & 47.94 & 29.64 & 57.21\\
95\% & 18.80 & 46.18 & 9.86 & 56.50\\
\bottomrule
\end{tabular}
\end{table}

Step-dLLM is both more accurate and faster than SparseD throughout this sweep. Near dense accuracy, Step-dLLM at $70\%$ sparsity takes 48.50 ms/step with 80.74\% accuracy, while SparseD at $50\%$ takes 59.94 ms/step with 80.59\% accuracy ($59.94/48.50\simeq1.24\times$). Relative to dense, the corresponding speedup is only $50.39/48.50\simeq1.04\times$. At $50\%$, Step-dLLM is slightly slower than dense. Sparse access must gather selected K/V blocks rather than stream a uniform dense layout; varying per-tile block counts also introduce load imbalance. Mask construction adds amortized overhead at anchor steps. With only a $2\times$ ideal reduction in attention FLOPs, these costs can offset savings, especially for short sequences.

Accuracy occasionally exceeds the dense baseline (e.g., 82.34\% at $60\%$ versus 81.04\% dense). Pruning weak or irrelevant connections is one possible explanation, but these single reported evaluations do not distinguish that hypothesis from evaluation variability. Small accuracy differences relative to dense should be interpreted conservatively. The sweep establishes one practical operating point, rather than generality across backbones, tasks, and lengths.

\FloatBarrier
\section{Periodic and Drift-Triggered Anchors}
\label{app:anchors}
We compare the periodic schedule with a key-representation drift trigger on GSM8K at $\rho=80\%$ and $S=256$. At every denoising step, the trigger computes the average cosine distance between the current key vectors and the keys stored at the last anchor. If this distance exceeds $\tau$, it performs full attention and rebuilds the mask, then updates the reference keys. Table~\ref{tab:drift} reports accuracy and the average number of mask refreshes per generation.

\begin{table}[ht]
\centering
\caption{Periodic versus key-drift-triggered anchors on GSM8K ($S=256$, $\rho=80\%$). Refresh count is an overhead indicator, not a measured latency ratio.}
\label{tab:drift}
\small
\begin{tabular}{lrr}
\toprule
Configuration & Accuracy (\%) & Avg. mask refreshes\\
\midrule
Periodic $f=0.2$ & 77.56 & 5\\
Adaptive key drift $\tau=0.02$ & 82.03 & 24\\
Adaptive key drift $\tau=0.001$ & 78.62 & 104.3\\
\bottomrule
\end{tabular}
\end{table}

The threshold $0.02$ improves accuracy by 4.47 percentage points, but requires $4.8\times$ as many refreshes. Lowering it to $0.001$ increases refreshes further while reducing accuracy, so more frequent triggering alone is not a monotonic predictor of quality. Every trigger invokes a dense anchor, and measuring drift adds work at each step. These results motivate the periodic schedule's deterministic refresh overhead; refresh counts quantify the number of dense anchors rather than a wall-clock latency ratio. Adaptive signals that are cheaper and better calibrated remain promising, especially for non-uniform decoding schedules. Alternating full attention and restricted reuse also appears in RefreshKV~\citep{xu2024refreshkv}; TeaCache~\citep{liu2025teacache} studies input-based cache decisions in video diffusion.

\FloatBarrier
\section{Saliency Pooling Comparison}
\label{app:pooling}
Table~\ref{tab:pooling} compares max pooling with average pooling using the supplementary accuracy differences. Max pooling preserves a block whenever it contains a strong query--key connection, whereas averaging emphasizes aggregate affinity. Max pooling is favorable in the higher-sparsity measurements, with the largest reported difference at $95\%$; average pooling is favorable at $50\%$ and $60\%$. Thus this design choice also depends on the operating regime. Quest~\citep{tang2024quest} and SeerAttention~\citep{gao2024seerattention} motivate examining extrema-based selection, while Native Sparse Attention~\citep{yuan2025nsa} provides a complementary compression-based approach. This table contains the max-versus-average comparison only.

\begin{table}[ht]
\centering
\caption{Reported accuracy difference between max and average pooling, in percentage points. Positive favors max pooling. These are differences, not absolute accuracy values.}
\label{tab:pooling}
\small
\setlength{\tabcolsep}{10pt}
\begin{tabular}{lrrrrrr}
\toprule
Sparsity & 50\% & 60\% & 70\% & 80\% & 90\% & 95\%\\
\midrule
Max $-$ average & $-2.43$ & $-0.46$ & $+0.76$ & $+1.06$ & $+0.83$ & $+3.64$\\
\bottomrule
\end{tabular}
\end{table}

\FloatBarrier
\section{Composition with Fast-dLLM Dual Caching}
\label{app:cache}
Sparse attention reduces the query--key interactions evaluated within a forward pass. Cache-based decoding reduces repeated computation and can advance several output tokens per step. We combine Step-dLLM with Fast-dLLM's dual cache~\citep{wu2025fastdllm} on GSM8K with LLaDA and batch size 1. Table~\ref{tab:cache} reports the separately measured recomputation and cache-composition results. Its $80\%$ recomputation accuracy (73.01\%) belongs to this integration experiment and should not be substituted for the 77.56\% in the main sweep.

\begin{table}[ht]
\centering
\caption{GSM8K accuracy (\%) and decoding progress with dual caching. Relative tokens/step is the reported multiple of single-token progress, not wall-clock speedup. The $0\%$ row uses the dense backbone.}
\label{tab:cache}
\small
\setlength{\tabcolsep}{6pt}
\begin{tabular}{rrrr}
\toprule
Sparsity & Recompute ($f=0.2$) & With dual cache & Relative tokens/step\\
\midrule
0\% & 81.04 & 81.43 & $3.80\times$\\
50\% & 81.65 & 80.06 & $3.51\times$\\
80\% & 73.01 & 70.74 & $2.71\times$\\
\bottomrule
\end{tabular}
\end{table}

At $50\%$ sparsity, dual caching reduces accuracy by 1.59 percentage points while reporting $3.51\times$ tokens per step. At $80\%$, the reduction is 2.27 points with $2.71\times$ tokens per step. This demonstrates compatibility in one integration setting, rather than a general wall-clock speedup or evidence that composition is lossless. Further evaluation with other cache methods and decoding schedules is left to future work.

\FloatBarrier
\section{Models, Benchmarks, and Baseline Settings}
\label{app:setup}
\paragraph{Models.} We evaluate pretrained LLaDA-1.5~\citep{zhu2025llada15} and Dream-7B-Instruct~\citep{ye2025dream} without additional training. Step-dLLM changes the attention computation and refreshes the sparse pattern; non-anchor steps recompute their query, key, and value representations. The attention block size is $b=16$ unless explicitly ablated; the semi-autoregressive generation block length is $B=32$, which is a different parameter. The default refresh interval ratio is $f=0.2$, with integer anchor spacing required in implementation.

\paragraph{Short-form benchmarks.} GSM8K evaluates grade-school mathematical reasoning, MATH-500 evaluates more challenging mathematical problem solving, and HumanEval evaluates code generation. We report the accuracy values recorded for the corresponding evaluations.

\paragraph{Long-context benchmarks.} NIAH evaluates retrieval with single and multiple needles, keys, queries, and values. RULER additionally includes variable tracking, word extraction, and question answering. Appendix~\ref{app:detailed-tables} lists the per-sub-task results. Their arithmetic means give the main NIAH and RULER columns. Long-context task difficulty and backbone behavior differ, so a single averaged column can hide sub-task exceptions.

\paragraph{Dense and StreamingLLM.} Dense attention provides the unpruned backbone reference. StreamingLLM~\citep{xiao2023efficient} retains sink tokens plus a sliding context window; our setup assigns sink tokens 10\% of the total generation length and derives the remaining window from the sparsity budget. This causal-decoder prior is a useful contrast but is not designed to track bidirectional attention throughout denoising.

\paragraph{SparseD.} We use the same $(b,B,\rho)$ as Step-dLLM and an initial full-attention fraction of 5\%. SparseD constructs its static sparse mask after this warm-up and reuses it. Its block-sparse stage uses \texttt{flex\_attention}, compiled into Triton kernels by TorchInductor. The refresh-interval ablation separately adds periodic refresh to this uniform-head-budget baseline; those rows must not be interpreted as the unchanged original SparseD algorithm.

\paragraph{Budgets and compute.} All reported experiments use a minimum allocation of one key block per head and query block. With $K$ the pre-floor top-$K$ budget, the retained-block count is $K+\sum_h\max(0,1-c_h(p))$ at query block $p$; the minimum allocation therefore relaxes nominal sparsity whenever a head wins no entries. All experiments use a single NVIDIA H100 GPU.

\FloatBarrier

\end{document}